# Dual-Axis RCCL: Representation-Complete Convergent Learning for Organic Chemical Space

Dejun Hu[1], Zhiming Li[1,3,*], Jia-Rui Shen[1,2], Jia-Ning Tu[1], Zi-Hao Ye[1], Junliang Zhang[1,3,4,5,*]

1 Department of Chemistry, Fudan University, Shanghai 200438, China

2 Shanghai Innovation Institute, Shanghai 200231, China

3 State Key Laboratory of Green Chemical Synthesis and Conversion, Department of Chemistry, Fudan University, Shanghai 200438, China

4 State Key Laboratory of Organometallic Chemistry, Shanghai Institute of Organic Chemistry, Chinese Academy of Sciences, Shanghai 200032, China

5 School of Chemistry and Materials, Yangzhou University, Yangzhou, Jiangsu 225002, China

*Email: zmli@fudan.edu.cn, junliangzhang@fudan.edu.cn

**Abstract:** Machine learning is profoundly reshaping molecular and materials modeling; however, given the vast scale of chemical space ($10^{30}$–$10^{60}$), it remains an open scientific question whether models can achieve convergent learning across this space. We introduce a Dual-Axis Representation-Complete Convergent Learning (RCCL) strategy, enabled by a molecular representation that integrates graph convolutional network (GCN) encoding of local valence environments, grounded in modern valence bond theory, together with no-bridge graph (NBG) encoding of ring/cage topologies, providing a quantitative measure of chemical-space coverage. This framework formalizes representation completeness, establishing a principled basis for constructing datasets that support convergent learning for large models. Guided by this RCCL framework, we develop the FD25 dataset, systematically covering 13,302 local valence units and 165,726 ring/cage topologies, achieving near-complete combinatorial coverage of organic molecules with H/C/N/O/F elements. Graph neural networks trained on FD25 exhibit representation-complete convergent learning and strong out-of-distribution generalization, with an overall prediction error of approximately 1.0 kcal/mol MAE across external benchmarks. Our results establish a quantitative link between molecular representation, structural completeness, and model generalization, providing a foundation for interpretable, transferable, and data-efficient molecular intelligence.

Graph neural networks (GNNs) and large language models (LLMs) are transforming molecular and materials modeling[1–4], enabling breakthroughs in reaction optimization[5–8], drug design[9–12] and materials discovery[13–18]. However, chemical space[19–22] is a high-dimensional, discrete domain of immense scale, commonly estimated to span $10^{30}$–$10^{60}$ possible molecules. This vastness poses a fundamental challenge that goes beyond model expressivity or training efficiency: whether machine learning models can achieve convergent learning with respect to chemical-space expansion[23]. Here, convergent learning does not refer to optimization convergence during training, but to convergence of model behavior as the underlying chemical space is systematically enlarged. Achieving this form of convergence requires not only sufficient model expressivity, but also that chemical structures be systematically coverable by a finite and interpretable representation. Developing molecular representations that enable systematic coverage and convergent learning thus poses a fundamental open problem in molecular science[24,25].

Prevailing graph-based enumeration approaches attempt to explore chemical space by systematically generating molecular graphs under valence and connectivity constraints, but inevitably suffer from exponential explosion, preventing systematic coverage and practical convergence[26,27]. The GDB (Generated Database) series exemplifies this limitation: GDB9, GDB11 and GDB17 enumerate 130k, 2.6M and 166B molecules with 9, 11 and 17 non-hydrogen atoms, respectively[28–30], highlighting the intrinsically non-convergent nature of chemical space under direct enumeration (Fig. 1a). Sampling-based strategies[31] alleviate the need for full enumeration and enable representative sampling within predefined chemical spaces, but do not directly address whether chemical space admits a finite, representation-complete basis required for convergent learning under systematic space expansion. Existing quantum-chemical datasets[32–37], including QM series[28,38–40] and PubChemQC series[41–44], often sampled from GDB[45] or experimental repositories such as PubChem[46,47] and ChEMBL[48,49], remain limited by biased structural coverage and lack a theoretical framework linking representation completeness to model generalization. As a result, convergent learning over chemical space remains an open problem.

At the electronic-structure level, molecular behavior is governed by the Schrödinger equation[50]. Modern valence-bond (VB) theory builds on this foundation by introducing orbital localization constraints to obtain localized bonding wavefunctions (Eq. 1)[51,52]. This localization enables a chemically interpretable description of σ/π bonding and non-bonded interactions, and can provide atom-wise, occupation-weighted orbital energies (Eq. 2) that characterize each atom's local valence environment. Graph Convolutional Networks (GCNs)[53] provide another local structure representation through neighborhood aggregation. The 0-hop representation (GCN0), corresponding to an atom's self-feature, captures its local valence coordination and aligns physically with atom-wise weighted orbital energies, making it a suitable descriptor of local valence environments. Definitions of all symbols are provided in Supporting Information Section 1.

$$\Psi_{VB} = \sum_k c_k \mathcal{A}\left[\phi_k^{(1)}(1)\phi_k^{(2)}(2)\cdots\phi_k^{(N_e)}(N_e)\chi_k\right] \quad (1)$$

$$E_{GCN0} = \frac{\sum_i o_i * e_i}{\sum_i o_i} \quad (2)$$

However, GCNs rely on local neighborhood aggregation and cannot adequately capture ring/cage topologies. To address this limitation, we introduce the No-Bridge Graph (NBG) representation[54], which explicitly encodes topological units whose properties, such as valence-bond cooperativity, conjugation and spatial strain, cannot be inferred from local neighborhoods alone. Because the combinatorial space of local atomic environments and ring/cage topologies are both relatively limited, the dual-axis representation combining GCN and NBG enables near-complete coverage of chemical space with finite data, supporting interpretable and convergent dataset construction (Fig. 1b).

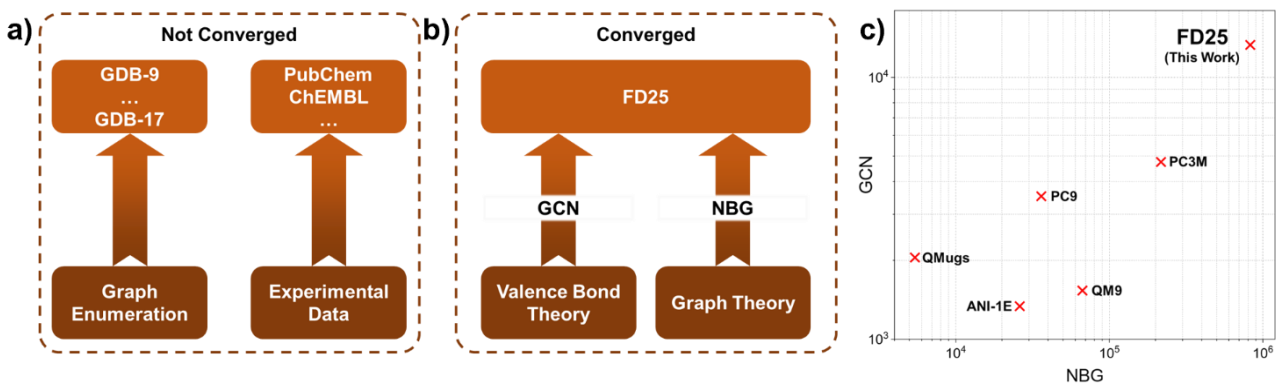

**Fig. 1** The Dual-Axis RCCL framework: from non-convergent to representation-complete convergent learning. a) Non-convergent chemical space sampling strategies based on graph enumeration (GDB-9–17) and experimental database sampling (PubChem, ChEMBL); b) Dual-Axis RCCL strategy integrating GCN (valence-bond local

environments) and NBG (ring/cage topologies); c) FD25 achieves substantially higher coverage than existing benchmark datasets across both the GCN and NBG dimensions.

Chemical space further involves long-range electronic delocalization, many-body polarization, and conformation-dependent cooperative effects[55], which cannot be fully captured by simple representations alone. When representation completeness is ensured, deep learning models can provide the required expressive capacity: GNNs encode distance-dependent interactions through multi-layer message passing, while LLMs capture higher-order electronic correlations via deep attention mechanisms. Within this context, the dual-axis GCN–NBG representation and deep neural networks enable interpretable and convergent learning over organic chemical space, a paradigm we term Dual-Axis Representation-Complete Convergent Learning (Dual-Axis RCCL).

Based on the Dual-Axis RCCL strategy, we constructed the FD25 dataset (Fudan University, 2025), comprising 2.1 million M06-2X/cc-pVDZ–optimized organic molecules containing H, C, N, O, and F. FD25 provides near-complete coverage of organic chemical space through 13,302 local valence units (GCN axis) and 165,726 ring/cage topology units (NBG axis), thereby achieving representation-complete coverage of the combined local-valence and ring/cage topological environment space (Fig. 1c). This coverage substantially exceeds that of existing benchmark datasets. Models trained on FD25 exhibit representation-complete convergent learning and strong cross-dataset generalization. The following sections detail the theoretical framework, dataset construction, and benchmark evaluation.

## Results

This section first establishes the theoretical basis of the Dual-Axis RCCL framework. We then define GCN encoding for local valence environments and NBG representation for ring/cage topologies, and use this foundation to refine chemical-space completion and enhance its diversity.

### *Design Principles*

To establish a representation-complete basis for organic chemical space, we first examined how local electronic environments respond to structural variations across different length scales, using atom-localized natural bond orbitals (NBOs)[56–58] as chemically interpretable probes. Aggregating core, valence ($\sigma_1$–$\sigma_4$), and antibonding orbital energies through an occupation-weighted scheme yields a single atom-local descriptor, $E_{GCN0}$ (Eq. 2). Using a dataset of 1,182 molecules containing tertiary carbon atoms (T1), we analyzed the statistical behavior of individual valence NBO energies and their $E_{GCN0}$ (Fig. 2b). Compared with individual σ-type NBOs, $E_{GCN0}$ exhibits substantially reduced fluctuations, with the standard deviation decreasing from 0.077–0.121 to 0.061 a.u.. This demonstrates that atomic-level energy aggregation enhances overall energetic stability by reducing sensitivity to incidental local perturbations. Consequently, $E_{GCN0}$ serves as a stable yet chemically meaningful descriptor of local valence environments and provides a robust basis for chemical space construction and model training.

Most variations in $E_{GCN0}$ arise from first-shell structural modifications. When the first-neighbor configuration of a tertiary carbon atom is fixed to methyl, methylene, and alkenyl (=CH) groups (148 molecules, T2 dataset), the distributions of both σ-orbital energies and $E_{GCN0}$ decrease markedly, with the standard deviation of $E_{GCN0}$ reduced to 0.020 a.u. (Fig. 2c). In this regime, the influence of more distant structural modifications becomes minimal. This demonstrates that first-shell valence environments form the minimal and sufficient units of local chemical diversity, motivating the central role of GCN1-level representations.

Beyond these local effects, ring/cage topologies induce systematic and reproducible shifts in $E_{GCN0}$ that cannot be captured by local neighborhood encoding alone. In the T1 dataset, tertiary carbon atoms in ring and chain environments show distinct average $E_{GCN0}$ values (−2.639 vs −2.665 a.u.), with corresponding standard deviations of 0.062 and 0.055 a.u.. Although these differences diminish once the first-shell structure is fixed, their reproducibility highlights the necessity of introducing an independent topological axis. Together, the dominance of first-shell valence environments and the irreducible contribution of ring/cage structures motivate the dual-axis GCN–NBG representation underlying the RCCL framework.

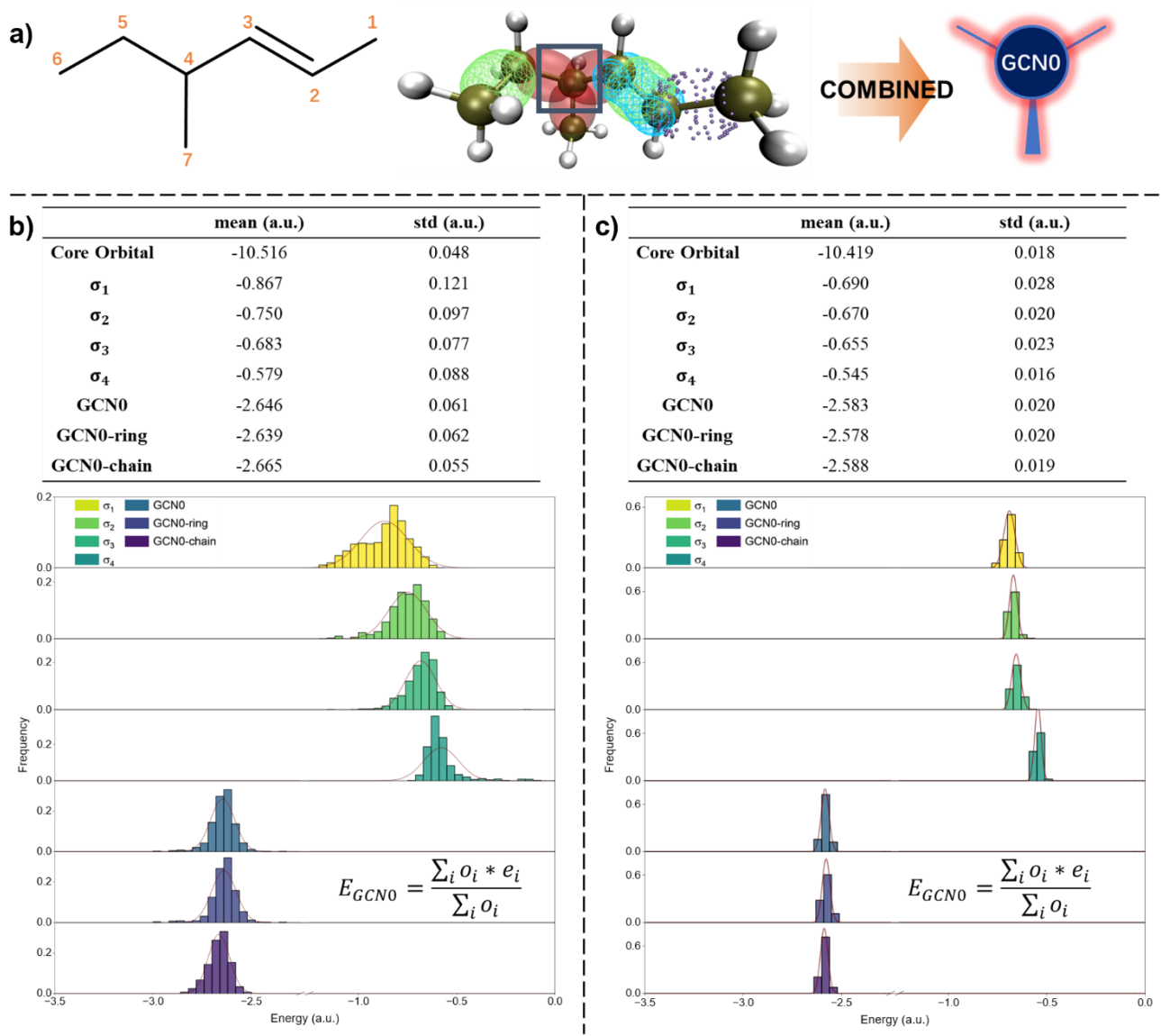

**Fig. 2** Sensitivity analysis of localized NBO energies in tertiary carbon environments. a) Schematic illustration of the representative C4 atom in 4-methylhex-2-ene and the NBO-based construction of the GCN0 representation. Individual core, valence ($\sigma_1$–$\sigma_4$), and antibonding orbitals are combined using an occupation-weighted averaging scheme to yield a single atom-local descriptor. b) Statistical distributions of valence σ-orbital ($\sigma_1$–$\sigma_4$) energies and $E_{GCN0}$, $E_{GCN0-ring}$, and $E_{GCN0-chain}$ for T1 dataset. Compared with individual σ orbitals, $E_{GCN0}$ show reduced fluctuations across different local structural environments. Systematic energy shifts are observed between ring and chain environments. c) Corresponding analysis for T2 dataset, in which the first-shell bonding environment of the central tertiary carbon atom is fixed to methyl, methylene, and alkenyl (=CH) groups. Under this constraint, variations arising from more distant molecular motifs lead to minimal changes in both individual and $E_{GCN0}$, and the separation between ring and chain environments is largely diminished.

### *Local Valence Environment Encoding: GCN*

To encode local atomic valence environments, we adopt a hierarchical graph-convolution representation (GCN0, GCN1, and GCN2) that captures increasingly extended local neighborhoods (0–2 hop; Fig. 3). GCN0 corresponds to an atom-centered valence descriptor defined by element type and first-shell coordination (non-hydrogen and hydrogen neighbors), aligning with localized NBO-derived electronic environments while avoiding uncertainty arising from explicit bond-order assignments. To ensure completeness of the local structural encoding, we systematically enumerate all singlet-state GCN0 configurations accessible to molecules composed of H, C, N, O and

F, yielding 30 distinct encoding types (Table S1). Extended Data Table 1 highlights GCN0 types that distinguish FD25 from existing benchmark datasets.

GCN1 was constructed by extending the GCN0 representation to include all 1-hop neighbors of the central atom, and served as the minimal representation that achieves structural completeness for localized valence environments (Fig. 3c). Each GCN1 code corresponds to the smallest molecule that can realize the associated local structure. Extended Data Fig. 1a presents the ten most frequent GCN1 types unique to FD25 dataset, together with their representative molecules.

GCN2 was generated by extending the GCN1 encoding by one additional level, allowing broader local structural correlations associated with potential substitution sites to be represented. This expansion enables the representation of higher-order geometric features such as inductive effects, p–π conjugation, and dihedral-angle variations. Owing to its rapidly expanding combinatorial space, GCN2 is employed to enhance in informational coverage rather than theoretical completeness. Together, the GCN hierarchy forms the local structural axis of the RCCL framework and complements global topological representations in downstream learning models.

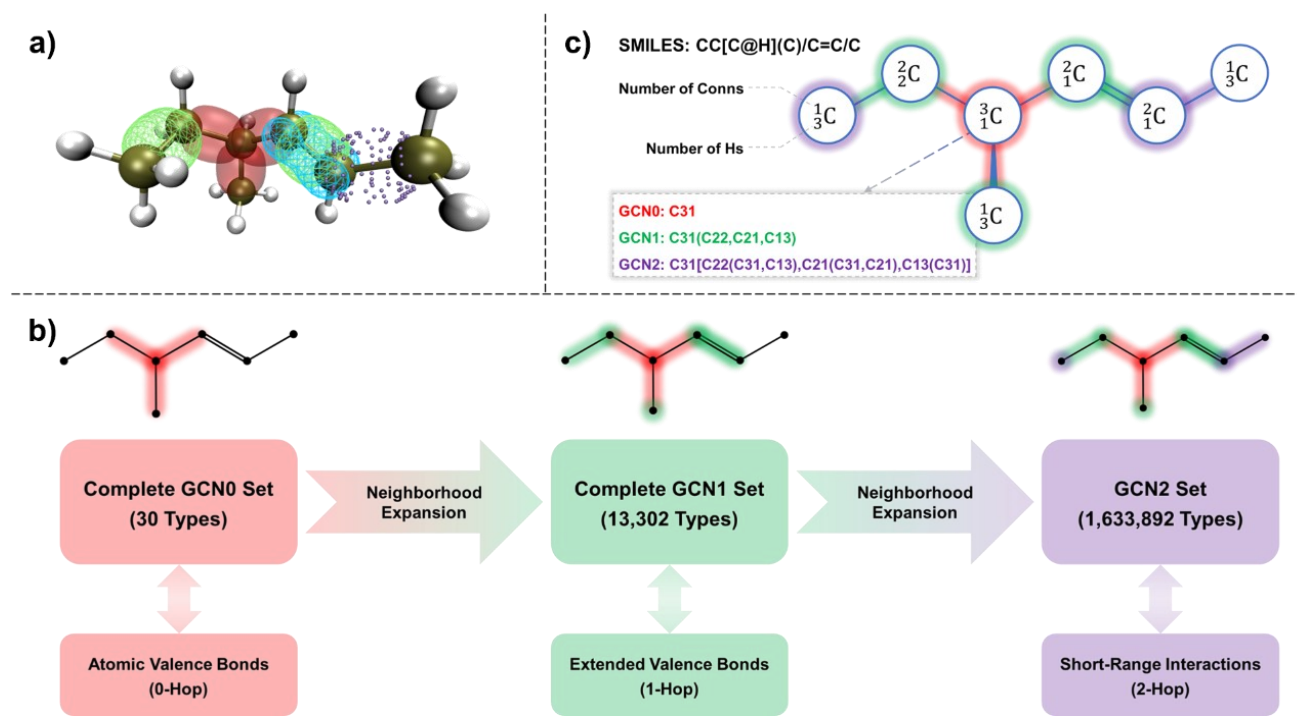

**Fig. 3** Hierarchical GCN encoding of local valence environments. a) Schematic illustration of NBO bonding orbitals in the carbon framework of 4-methylhex-2-ene; b) Conceptual representation of the hierarchical GCN encoding scheme, illustrating the progressive extension from GCN0 to GCN2; c) Representative GCN0, GCN1, and GCN2 encodings for the tertiary carbon atom in 4-methylhex-2-ene, demonstrating the progressive incorporation of first- and second-shell neighborhood information.

## *Topological Principle of NBG*

To systematically represent ring/cage topologies in organic chemical space, we introduce the NBG principle. As illustrated in Fig. 4a, cut vertices and cut edges (bridges) denote structural elements whose removal increases the number of connected components[59], denoted as $\mathcal{V}_c$ and $\mathcal{E}_c$. In molecular graphs composed of carbon and hydrogen atoms, ring and cage topologies that contain no bridges are classified as NBG0 (Fig. 4c). Such structures are indivisible in the graph-theoretical sense, as no single bond cleavage can generate two disconnected fragments. They can thus be regarded as fundamental topological units widely encountered in synthetic chemistry and exhibit theoretical structural completeness.

To establish a structurally complete NBG0 encoding space, we developed a simplified molecular graph–generation strategy (Fig. 4b), producing an NBG0 topological dataset containing 165,726 structures. We constrain the enumerated non-hydrogen atom count (Na) to no more than 13, thereby maintaining a balance between computational efficiency and structural coverage. Compared with the 2,482 NBG0 structures in PC3M, the theoretical enumeration used here expands the coverage by two orders of magnitude, providing a systematic foundation for studying ring/cage effects in organic chemical space. Extended Data Fig. 2 shows the ten most frequent and unique NBG0 types in FD25 dataset.

To further enhance topological diversity and capture substituent–ring/cage cooperative effects (e.g., the hydroxyl–aromatic interaction in phenol), we extended the NBG0 framework to construct the NBG-Plus space through controlled topological modifications (Fig. 4c), such as isovalent substitution of carbon atoms with heteroatoms (N, O), and cut vertices and bridges expansion. Molecules such as benzoic acid and 4-(1-phenylethyl)pyridine exemplify typical NBG-Plus structures. NBG-Plus represents a high-dimensional extension of NBG0, introducing both local and nonlocal effects through controlled substitution and bridging operations, while constraining higher-order interactions to a three-body approximation to balance model efficiency and interpretability.

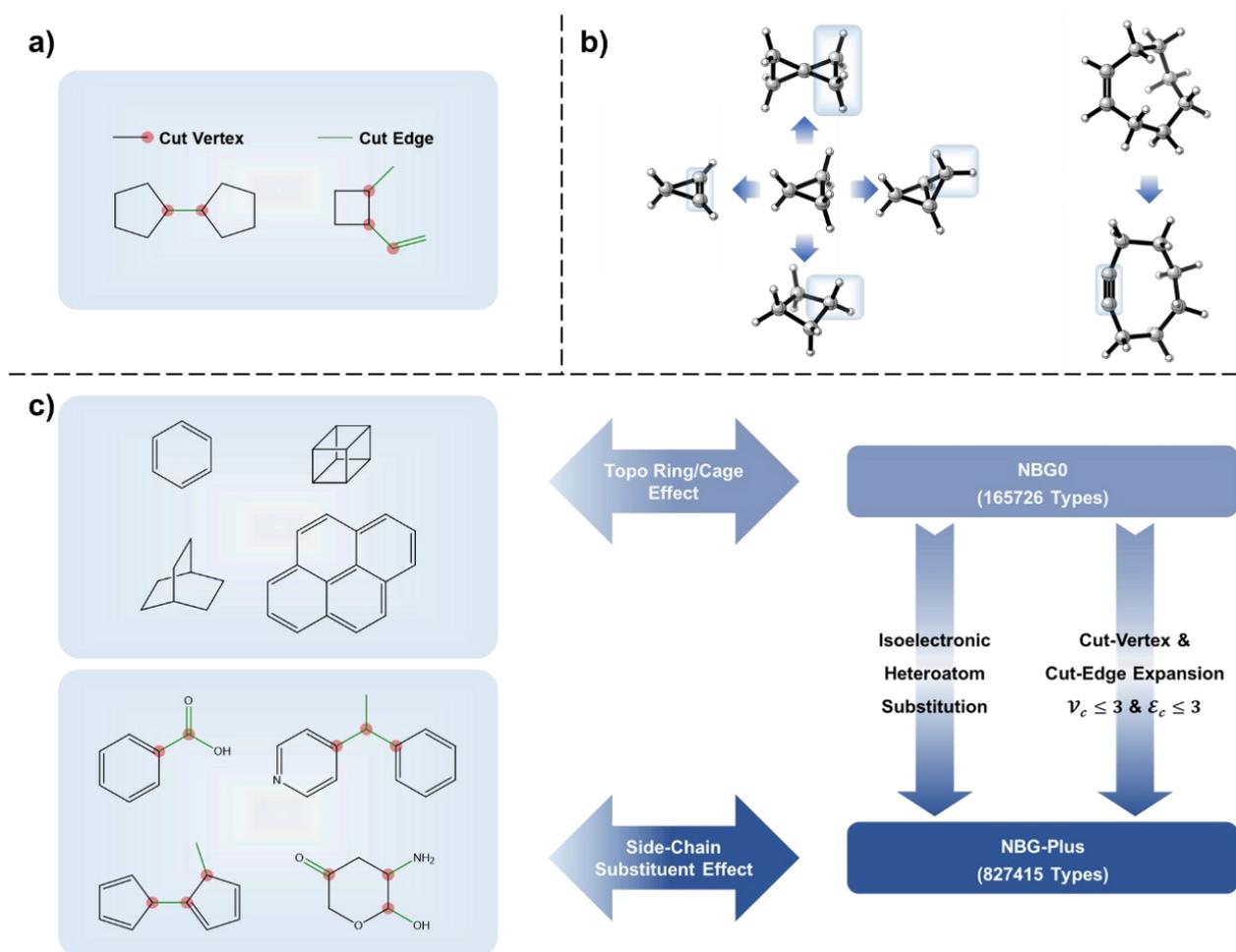

**Fig. 4** Topological principle and hierarchical construction of NBG representations. a) Definition of cut vertices and cut edges, where red nodes represent cut vertices and green edges denote cut edges; b) Schematic illustration of the NBG0 representation, highlighting ring and cage topologies without bridges as fundamental topological units; c) Representative NBG0 and NBG-Plus structures, illustrating how ring and cage frameworks are systematically extended through controlled substitution and connectivity augmentation to capture cooperative topological effects.

Conceptually, NBG-Plus is related to the scaffold paradigm[60–64] in medicinal chemistry. However, unlike conventional scaffolds, NBG-Plus retains minimal partition graphs that form the structural basis for functional-group combinations and conformational variations. Isovalent derivatives of NBG0 can be further decomposed along cut edges to yield minimal molecular fragments align with the fragmentation paradigm in chemistry. This approach enables the construction of a structurally complete topological space at substantially reduced computational cost. Together, NBG0 and NBG-Plus define the ring/cage topological axis of the RCCL framework, complementary to the local valence representations encoded by GCN.

### *Chemical Space Completion and Diversity Enhancement*

Building upon the GCN and NBG representation principles, the chemical space was further completed and diversified along three complementary dimensions. First, systematically constructed small-molecule chemical space ($N_a \leq 6$) was established to capture all constitutional and conformational variations, including high-energy local minima that are sparsely represented in existing datasets. Because the structural and conformational variations of larger molecules are largely governed by local bonding environments, this $N_a \leq 6$ chemical space provides a foundational basis for constructing extended and continuous chemical spaces. Second, a global-minimum subset was constructed to provide high-coverage reference structures for GNN and LLM models in minimum-energy conformation prediction and related property modeling. Third, an elemental-combination uniformity principle was introduced to balance atomic-composition distributions across organic chemical space and mitigate size-induced biases. Together, these design choices ensure both coverage and diversity, enabling robust and transferable molecular representations.

### *Construction of FD25*

Based on the GCN encoding principle, the NBG topological framework, and the chemical-space completion and diversity enhancement strategy, we constructed the core dataset of FD25 (Fig. 5a) as a systematic molecular representation space for chemical-space expansion, which comprises 1,612,929 unique molecules and 2,015,740 conformations. To further evaluate model performance and generalization in GNN- and LLM- based learning tasks, an extended dataset of FD25 (Fig. 5a) was constructed by randomly sampling 650k medium-sized HCNOF molecules ($9 \leq N_a \leq 32$) from the PubChem database (version 2022.10). For subsequent evaluations, the extended dataset was evenly divided into three subsets, one combined with the core dataset for model training, and the remaining two used as validation and test sets, respectively.

Together, the core and extended datasets constitute the complete FD25 dataset. All molecular structures were optimized at the M06-2X/cc-pVDZ level. For each optimized structure, ten quantum-chemical properties were collected, including total energy, enthalpy, Gibbs free energy, vibrational frequency, HOMO energy, LUMO energy, molecular volume, polarizability, solvation energy, and atomic charges.

### *FD25 Dataset Characterization*

To evaluate the effectiveness of the dual-axis expansion strategy, we compared the FD25 dataset with four representative quantum chemistry datasets, PC3M, QM9, PC9, and ANI-1E (Fig. 5b, Supporting Information Section 4). These datasets were selected based on their consistent use of DFT-level geometry optimization, availability of quantum chemical properties, and representative coverage across molecular scale and chemical space. In terms of molecular scale (Number of Mols), FD25 (2173k) is slightly smaller than PC3M (2520k), yet substantially larger than QM9, PC9, and ANI-1E. Notably, FD25 offers approximately 1309k global minimum structures, significantly enhancing its ability to characterize three-dimensional structural diversity.

Across the GCN hierarchy, FD25 exhibits the most complete GCN0 coverage among the five datasets. The benchmark datasets consistently miss GCN0 types corresponding to quaternary ammonium motifs and other high-energy local bonding environments, leading to gaps at the most basic level of chemical representation. This advantage extends to higher levels of the hierarchy. FD25 exhibits approximately 3-fold and 2.5-fold increases in the numbers of GCN1 and GCN2 types relative to PC3M, respectively. This improvement reflects the systematic construction of GCN1 space, enabling extensive coverage of bonding orbital environments despite a moderate molecular count. FD25 also exhibits substantially higher diversity in NBG0 and NBG-Plus metrics, exceeding PC3M by factors of about 67 and 3.8 times, respectively, indicating broader coverage of ring/cage motifs. A similar advantage is observed under the scaffold metric.

We further performed an overlap analysis with QM9 and PC3M (Fig. 5c) across two basic structural dimensions, GCN1 and NBG0. FD25 exhibits significantly broader coverage in GCN1 and NBG0 metrics, which emphasize local bonding patterns and ring/cage motifs, further highlighting its representational completeness. Additional statistical analyses of molecular size, binding energy, and scaffold topology (Fig. S1–S2) further confirm the broad coverage and structural diversity of FD25.

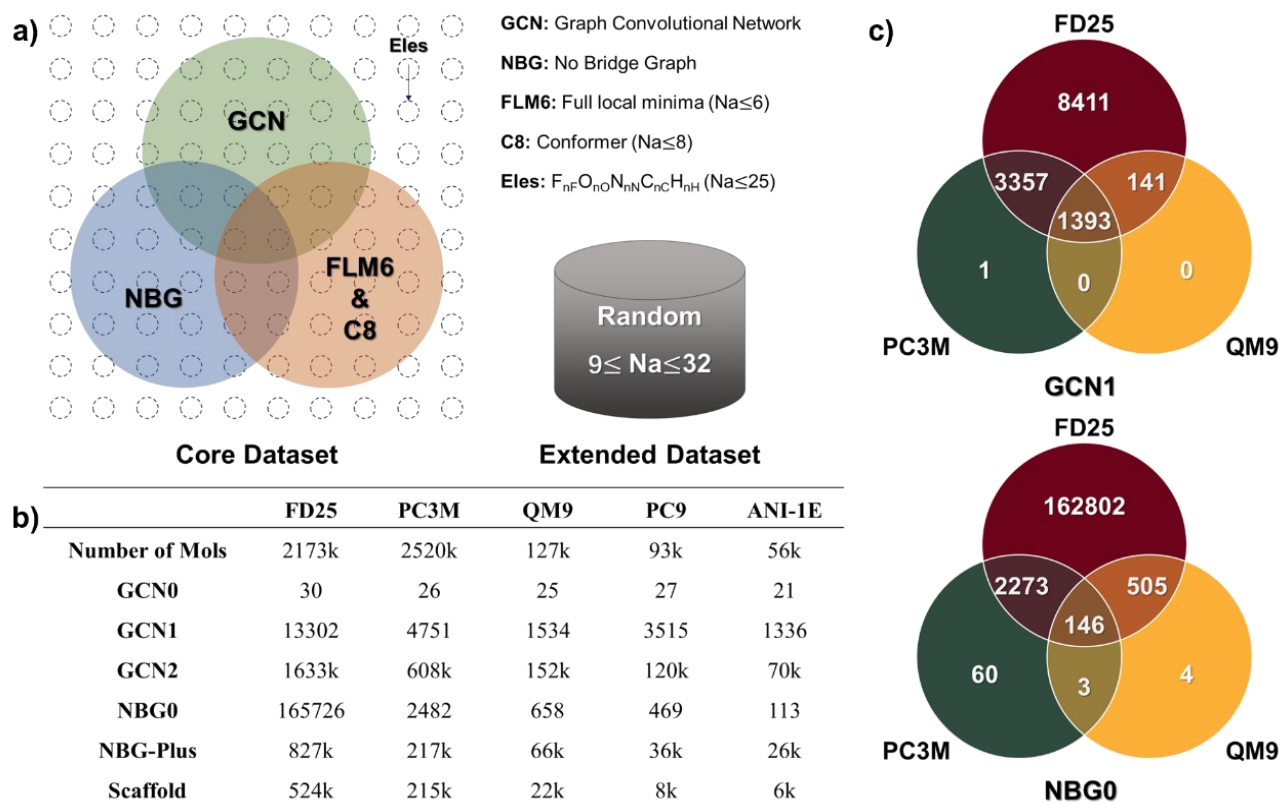

**Fig. 5** Composition and intersectional diversity analysis of FD25 dataset. a) Overall composition of FD25 dataset; b) Comparative statistics of FD25 and four representative quantum chemistry datasets; c) Overlap analysis of FD25 with QM9 and PC3M.

## Model Evaluation and Generalization Tests

We evaluated the modeling accuracy and generalization performance of FD25 on binding energy (Eq. S3) prediction tasks using GNN models. The evaluation comprised four components: convergence tests under structural expansion along the GCN2 and NBG-Plus axes, error attribution analysis across GCN1, NBG-Plus, and scaffold features, cross-dataset evaluations against QM9 and PC3M, and benchmark prediction across multiple quantum properties. Model architectures and training configurations are detailed in Supporting Information Section 6.

**Convergence of Structural Expansion**

Given that FD25 already achieves high coverage along the GCN1 and NBG0 axes, we focus on higher-complex structural dimensions (GCN2 and NBG-Plus) to evaluate convergence under structural expansion. Specifically, we monitored the evolution of binding energy prediction errors as progressively more complex structures were introduced, using error saturation as the criterion for convergent learning across organic chemical space.

To construct the structural expansion sequence, hierarchical complexity levels were defined for GCN2 and NBG-Plus. GCN2 was divided into 17 levels (Level 0–16) based on the maximum GCN2 value of neighboring non-hydrogen atoms (Na), while NBG-Plus was divided into 23 levels (Level 0–22) according to the maximum number of cut vertices and bridges. Detailed level definitions are provided in Table S6. Starting from $Level_{GCN2}=6$ and $Level_{NBG\text{-}Plus}=1$, five progressively expanded subsets (step0-step4; Extended Data Table 2) were constructed with

Step4 corresponding to the full FD25 dataset. Training and validation sets were dynamically updated at each step, while the test set remained fixed (see Supporting Information Section 6.3) to ensure comparability.

Convergence-threshold analysis (Fig. 6a) shows that test errors are high at the initial expansion stage (step 0, MAE > 10 kcal/mol). When the structural space is expanded to step 2 (Level$_{GCN2}$=8, Level$_{NBG\text{-}Plus}$=3; ~1.3M molecules), the error rapidly decreases into the chemical-accuracy regime (~1 kcal/mol). Further expansion to step 3 (Level$_{GCN2}$=9, Level$_{NBG\text{-}Plus}$=5; ~1.7M molecules) yields diminishing returns, indicating near-convergent performance. Notably, FD25 extends to Level$_{GCN2}$=16 and Level$_{NBG\text{-}Plus}$=22 (step 4), far beyond the structural complexity required for model convergence, with no benefit from additional higher-order structures.

These results demonstrate a well-defined convergence threshold in organic chemical space under representation completeness. Consequently, the Dual-Axis RCCL strategy enables data-efficient organic chemical-space coverage: with a nearly complete basis at GCN1 and NBG0, modest extensions to GCN2 and NBG-Plus capture essential chemical diversity while substantially reducing dataset size and training cost. Compared with GDB-11, a graph-theoretic enumerated dataset with 26.4 million molecules, FD25 contains only 2.23 million molecules, yet provides sufficient chemical-space coverage to support large-model training.

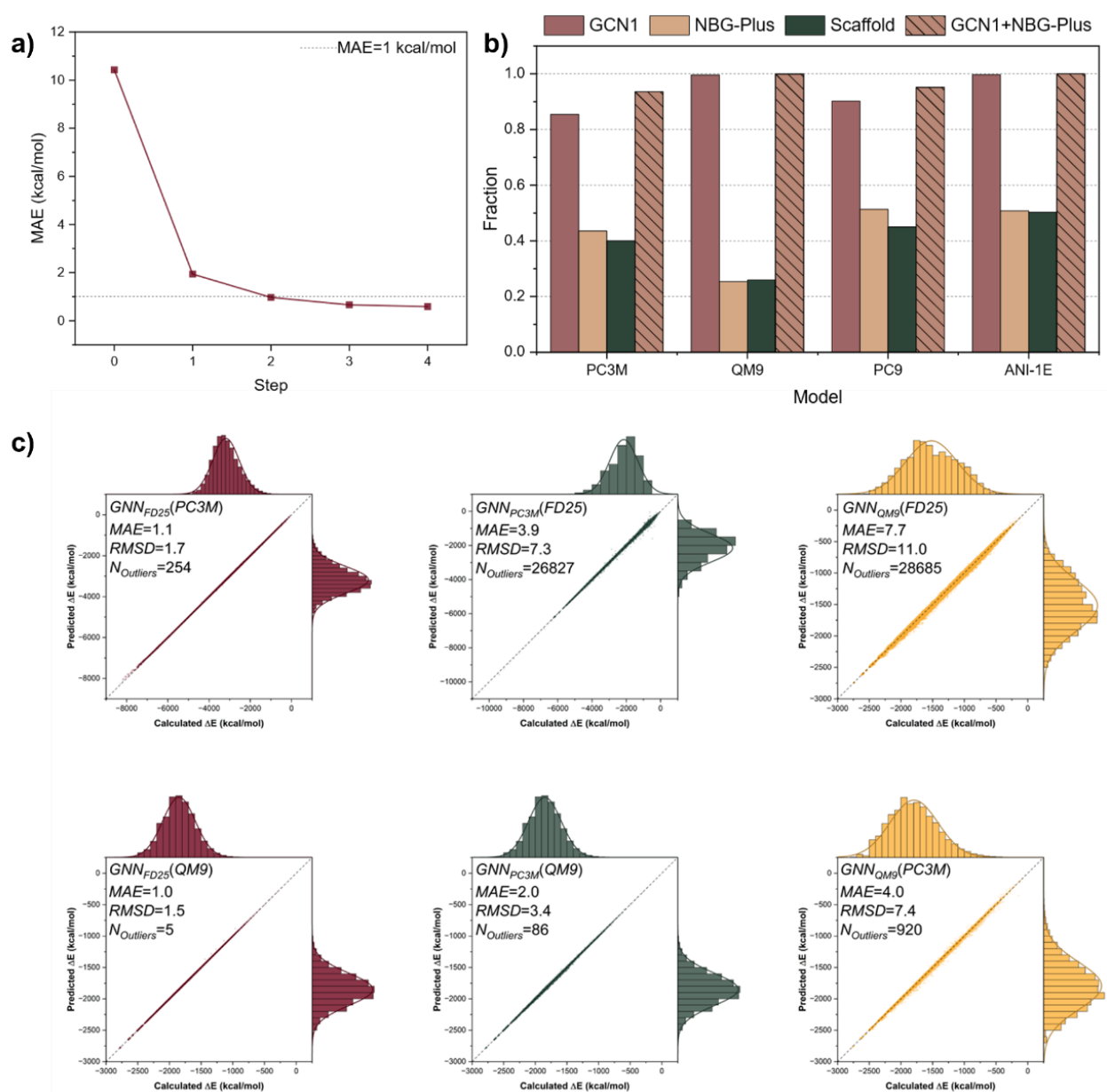

**Fig. 6** Structural expansion convergence and cross-dataset generalization analysis. (a) Convergence of binding-energy prediction error under progressive structural expansion along the GCN2 and NBG-Plus axes; (b) Relative contributions of GCN1, NBG-Plus, scaffold, and their combination to cross-dataset generalization.; (c) Cross-dataset prediction performance shown by parity plots and error distributions for $GNN_{PC3M}$, $GNN_{QM9}$, $GNN_{FD25}$, together with outlier statistics. MAE and RMSD are reported in kcal/mol. $N_{outliers}$ denotes the number of molecules with absolute prediction errors exceeding 30 kcal/mol.

**Error Attribution Analysis**

The structural expansion-convergence tests above demonstrate that increasing structural complexity, either through atomic environment (GCN2) or topological motifs (NBG-Plus), systematically reduces prediction errors and enhance model performance. However, these structural dimensions are not orthogonal; they are often strongly coupled, complicating direct attribution of model errors. To decouple these effects, we trained four independent GNN models on PC3M, QM9, PC9 and ANI-1E, denoted as $GNN_{PC3M}$, $GNN_{QM9}$, $GNN_{PC9}$, and $GNN_{ANI-1E}$, and evaluated all models on FD25 for systematic error attribution analysis.

We constructed three "structure-complement subsets" (GCN1-Comp, NBG-Plus-Comp, and Scaffold-Comp), each comprising FD25 molecules that contain a given structural feature absent from the corresponding training dataset. To ensure statistical robustness, we selected the top 20,000 FD25 molecules with the largest prediction errors for each model, while restricting Na to the coverage range of the training dataset to control for energy-distribution effects. For instance, QM9 covers Na values between 1 and 9; thus, only FD25 molecules with Na≤9 were selected for evaluation in that case.

As shown in Fig. 6b, highly consistent attribution patterns are observed across all four training datasets. Among the top 20,000 error samples, GCN1-Comp accounts for 85–99%, and NBG-Plus-Comp for 25–50%, with the two together explaining 94–100% of the error cases. The dominance of GCN1 remains stable across different Top-k thresholds (500–20,000; Fig. S4), indicating that its impact on prediction error is invariant to specific error cutoffs. We attribute this to the fact that GCN1 encodes the first-shell atomic environment, serving as a primary input to graph-based chemical learning, and thus acts as the most error-sensitive structural descriptor. Notably, NBG-Plus and scaffold show similar contributions, implying that they offer functionally related representations that complement GCN1. These findings provide a structural basis for the cross-dataset generalization analysis discussed below.

**Cross-Dataset Testing**

To further evaluate the impact of training data distribution on cross-dataset generalization, we additionally trained a GNN model on FD25 ($GNN_{FD25}$). Together with the previously trained $GNN_{PC3M}$ and $GNN_{QM9}$, each model was then evaluated on the remaining two datasets. This design enables a systematic assessment of how structural coverage differences influence predictive performance, thereby revealing the extent to which missing structural motifs constrain model generalization.

Due to variations in DFT protocols across datasets, direct comparison of raw energies is confounded by methodological inconsistencies and structural shifts. To mitigate such systematic discrepancies, we project all energies onto a unified reference space using linear regression. Beyond conventional normalization based on molecular or atomic composition (Eq.3-4), we incorporate GCN1 structural descriptors (Eq. 5), expressing molecular binding energies as a linear combination of 13,302 GCN1 atom types:

$$E_1 = c_0 * E_0 + b_0 \tag{3}$$

$$E_1 = c_0 * E_0 + \sum c_i * N_i + b_0 \tag{4}$$

$$E_1 = c_0 * E_0 + \sum c_i * N(GCN_1)_i + b_0 \tag{5}$$

Where $E_0$ and $E_1$, are properties computed under different DFT settings. $N_i$ and $N(GCN_1)_i$ denote atomic or GCN1-type counts, with $c_i$ and $b_0$ are fitted coefficients.

Cross-dataset testing (Fig. 6c) reveals that the decisive role of structural coverage. $GNN_{FD25}$ exhibits strong transferability, achieving MAEs of ~1.0 kcal/mol on PC3M and QM9, with outliers (absolute prediction errors > 30 kcal/mol) limited to 254 and 5 molecules, respectively. In contrast, $GNN_{PC3M}$ shows degraded performance on QM9 (MAE = 2.0 kcal/mol, $N_{outliers}$ = 86), which worsens substantially on FD25 (MAE = 3.9 kcal/mol, $N_{outliers}$ = 26,827), highlighting limited coverage of GCN1-type local environments. $GNN_{QM9}$ generalizes poorly, with MAEs of 7.7 and 4.0 kcal/mol on FD25 and PC3M, and $N_{outliers}$ reaching 28,685 and 920, respectively, reflecting insufficient structural diversity for extrapolation.

Taken together, $GNN_{FD25}$ maintains stable accuracy across multiple datasets with distinct distributions, demonstrating that comprehensive structural coverage is essential for robust generalization. These results establish FD25 as a structural benchmark for evaluating generalization capability of molecular models.

**Multi-Property Benchmarking of $GNN_{FD25}$**

We next benchmarked the predictive performance of $GNN_{FD25}$ across ten quantum chemical properties: total energy (E), enthalpy (H), free energy (G), HOMO and LUMO orbital energies, HOMO–LUMO gap, aqueous solvation energy ($E_{solv}$), its non-electrostatic solvation energy (SMDE), isotropic polarizability (α), and atomic Mulliken charges. These targets span five classes: (1) extensive thermodynamic energies (E, H, G); (2) frontier orbital energies (HOMO, LUMO, gap)[65]; (3) polarizability (α); (4) solvation energies ($E_{solv}$, SMDE); and (5) atomic partial charges (Mulliken), and an independent GNN was trained per property class to avoid multi-task interference.

Model performances are summarized in Extended Data Table 3. The GNN models achieves chemical accuracy for multiple targets: MAEs for E, H, and G were within 0.7–0.8 kcal/mol, substantially below the 1 kcal/mol benchmark. MAE for the HOMO–LUMO gap is 1.6 kcal/mol. Notably, solvation-related properties exhibit particularly strong performance, with MAEs of approximately 0.1 kcal/mol for both $E_{solv}$ and SMDE, demonstrating robust generalization.

**Discussion**

We present a Dual-Axis RCCL strategy that integrates local valence environments (GCN axis) and ring/cage topologies (NBG axis), providing a unified descriptor of organic chemical-space coverage and introducing the concept of representation completeness for the first time. With near-complete GCN–NBG representation and modest structural extensions, high-coverage molecular datasets across organic chemical space can be constructed. Coupled with sufficiently expressive GNNs or LLMs, RCCL thus enables convergent learning for organic molecular systems.

Using this strategy, we constructed the FD25 dataset, comprising 2.1 million M06-2X/cc-pVDZ validated HCNOF molecules. FD25 provides near-complete coverage of 13,302 GCN1 valence units and 165,726 NBG0 topological units, exceeding existing benchmark datasets. $GNN_{FD25}$ achieve representation-complete convergent learning with strong cross-dataset generalization, highlighting the central role of GCN1 environments in molecular prediction.

FD25 provides a reliable training and evaluation basis for building first-principles-faithful GNN and LLM models, and also can serve as a benchmark for machine-learning potentials. Because the dataset is generated through extensive dynamical sampling and geometry optimization and includes key DFT properties, such as free energies, solvation energies and other optimized-structure quantities, FD25 further enables models to directly predict DFT-level properties of optimal molecular structures without explicit 3D sampling. Future work will extend Dual-Axis RCCL beyond equilibrium HCNOF systems to multi-element, excited-state, and transition-state structures, and will support the design of new linear molecular representations for more efficient chemical-space encoding.

## Methods

### GCN encoding

A complete GCN1 encoding space was generated via Algorithm S1, and 47,870 configurations in total was generated. To guarantee the chemical validity of these fragments, all 47,870 GCN1 molecules were geometrically optimized at the M06-2X/cc-pVDZ level. After removing non-convergent cases, a benchmark subset of 13,302 stable local valence-bond structures was obtained. Notably, two special GCN1 structures found in the PC3M and PC9 datasets are absent here (Extended Data Fig. 1b), as their geometries failed to converge at the M06-2X/cc-pVDZ level. Constructing a full GCN2 set is computationally impractical (Algorithm S2). The dataset ultimately includes 1,633,892 distinct GCN2 types.

Both GCN1 and GCN2 are concatenated in reverse lexical order, and separated by commas, and enclosed in parentheses and square brackets, respectively.

### Construction of the NBG0 and NBG-Plus datasets

To establish a structurally complete NBG0 encoding space, we developed a simplified molecular graph–generation strategy, which involves three steps: (1) Starting from cyclopropane, five basic recursive operations are applied (Fig. 4b): ring expansion of methylene ($CH_2$) into cyclopropane, single-bond ring expansion into cyclopropane, single-bond methylene insertion, conversion of a single bond to a double bond, and conversion of a double bond to a triple bond. (2) From the $N_a \leq 6$ molecular space, NBG0 structures composed of C and H atoms were selected, recursively expanded using the same five operations, and refined through DFT optimization, merging, and filtering. (3) Typical large ring systems—including nested rings, fused rings, and partial spiro compounds—were further incorporated.

To construct the NBG-Plus dataset, NBG0 structures were further augmented through controlled topological modifications. First, isovalent substitution of carbon atoms with heteroatoms (N, O) was introduced to enrich the elemental diversity of ring and cage structures. Second, under the constraint of no more than three cut vertices and cut edges ($\mathcal{V}_c \leq 3 \& \mathcal{E}_c \leq 3$), specific side-chain substituents (e.g., –OH, –$NH_2$, =O) and bridging linkages were incorporated, enabling multiple ring/cage units to couple into more complex topological assemblies. These constraints limit higher-order interactions while preserving chemically relevant cooperative effects between substituents and cyclic cores. Typical NBG-Plus structures include benzoic acid (two cut vertices and three cut edges) and 4-(1-phenylethyl)pyridine (three cut vertices and three cut edges), as illustrated in Fig. 4c.

### Chemical Space Completion and Dataset Construction

Building upon the GCN and NBG construction principles, chemical space completeness and diversity were achieved by expanding along three complementary dimensions.

**(a) Systematic construction of the $N_a \leq 6$ molecular space**

We systematically generated a comprehensive constitutional and conformational chemical space of neutral singlet molecules containing up to six non-hydrogen atoms ($N_a \leq 6$). Through flexible bond scanning and local-minimum sampling, 151k unique molecules and 275k stable conformers were obtained. Molecules with up to six non-hydrogen atoms were extracted from the GCN and NBG datasets. Bonded pairs were subjected to extension flexible scans

(0.15 Å increments, extended to 3.0 Å), while non-bonded pairs were subjected to contraction scans (-0.15 Å increments, reduced to 1.0 Å). All local-minimum geometries identified during these scans were optimized at M06-2X/cc-pVDZ level and collected.

Compared with the PC3M dataset, which contains only 6,757 molecules in this size range, FD25 substantially expands coverage, particularly by including high-energy local minima. Because structural and conformational variations in larger molecules can often be decomposed into local combinations of six-atom fragments, the complete $N_a \leq 6$ space provides a foundational basis for constructing extended chemical spaces.

**(b) Global-minimum subset ($N_a \leq 8$)**

Considering that the local atomic environment required to describe chemical bonding typically involves up to eight atoms, we further collected complete conformational data for molecules with $6 < N_a \leq 8$ and identified the global minimum conformation for each. For selected larger molecules ($N_a > 8$), global-minimum conformations were also included. Combined with the $N_a \leq 6$ chemical space, these data form the FD25-Global-Minimum subset, comprising approximately 1.3 million entries, and provide a high-coverage dataset for minimum-energy conformation prediction and related property modeling.

**(c) Elemental-combination uniformity**

To enhance model generalization across atomic-composition space, we introduced the principle of elemental-combination uniformity. Within the range of $N_a \leq 25$, this principle enforces a balanced distribution of HCNOF elemental combinations, preventing over- or under-representation of specific compositions and mitigating size-related biases. For all molecular compositions of the general form $F_{n_F}O_{n_O}N_{n_N}C_{n_C}H_{n_H}$ ($n \geq 0$), random sampling was performed to ensure that each composition is represented by at least two distinct molecular structures.

## Data Availability

The FD25 dataset generated in this study is publicly available via Zenodo (DOI: 10.5281/zenodo.17910855). It comprises 2,668,346 organic molecules computed at the M06-2X/cc-pVDZ level of theory and includes molecular structures, quantum chemical properties, dataset splits, pre-trained model weights, and linear alignment coefficients. The released data contain atomic Mulliken charges, thermochemical and electronic properties, and implicit aqueous solvation energies computed using the SMD model. Accompanying resources include predefined training, validation and test splits (split.json); indices for the core subset (core_index.csv) and global minimum conformations (conf_info.csv); pre-trained GNN model weights for property prediction; and linear alignment coefficients for cross-dataset correction. All data files are provided with SHA-256 checksums to ensure data integrity.

## Code Availability

The code for GNN training and prediction is publicly available at: https://github.com/hdj020402/GNN. A frozen release corresponding to the experiments in this paper is provided at: https://github.com/hdj020402/GNN/releases/tag/fd25-v1.0.1-fixed-readme. This release contains the exact source code, dependencies, and documentation required to reproduce the results presented in this study. The main branch of the repository will continue to be updated and may differ from the version associated with this paper.

## Acknowledgement

We gratefully acknowledge the funding support of NSFC (22471042, 22031004, 22271053, 22471040), Science and Technology Commission of Shanghai Municipal (23ZR1404800), AI for Science Foundation of Fudan University (FudanX24AI024), Shanghai Municipal Education Commission (20212308), National Key R&D Program of China (2021YFF0701600). The computations in this research were performed using the CFFF platform of Fudan University. The authors also thank Prof. Zhenhua Li and Prof. Zhangjie Shi (Fudan University) for providing partial computational support.

## Author contributions

Z.L. and J.Z. conceived the study. D.H. developed the software and performed the formal analysis. D.H., J.S., J.T. and Z.Y. carried out the investigations. Z.L. and J.Z. acquired funding and provided resources. Z.L. and J.Z. supervised the project. D.H. and Z.L. wrote the original draft; all authors discussed the results and reviewed the manuscript.

## Competing interests

The authors declare no competing interests.

**Supplementary information** is available in the online version of the paper.

**Correspondence** and requests for materials should be addressed to Z.L.* (zmli@fudan.edu.cn)

**Extended Data Table 1 GCN0 types distinguishing FD25 from existing benchmark datasets.**

GCN0 types present in FD25 but absent from PC3M, PC9, QM9 or ANI-1E, shown with their central atom types, local coordination environments, and representative molecular examples.

| GCN0 Type | Representative Molecule | Absent from |
|---|---|---|
| C02 | 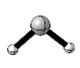 | PC3M, PC9, QM9, ANI-1E |
| C10 | 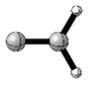 | QM9, ANI-1E |
| N01 | 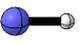 | PC3M, QM9, ANI-1E |
| N13 | 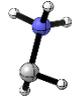 | ANI-1E |
| N22 | 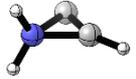 | PC3M, PC9, ANI-1E |
| N31 | 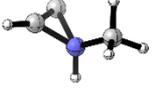 | PC3M, PC9, ANI-1E |
| N40 | 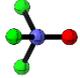 | QM9 |
| F01 | 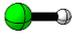 | QM9, ANI-1E |
| F10 | 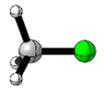 | ANI-1E |

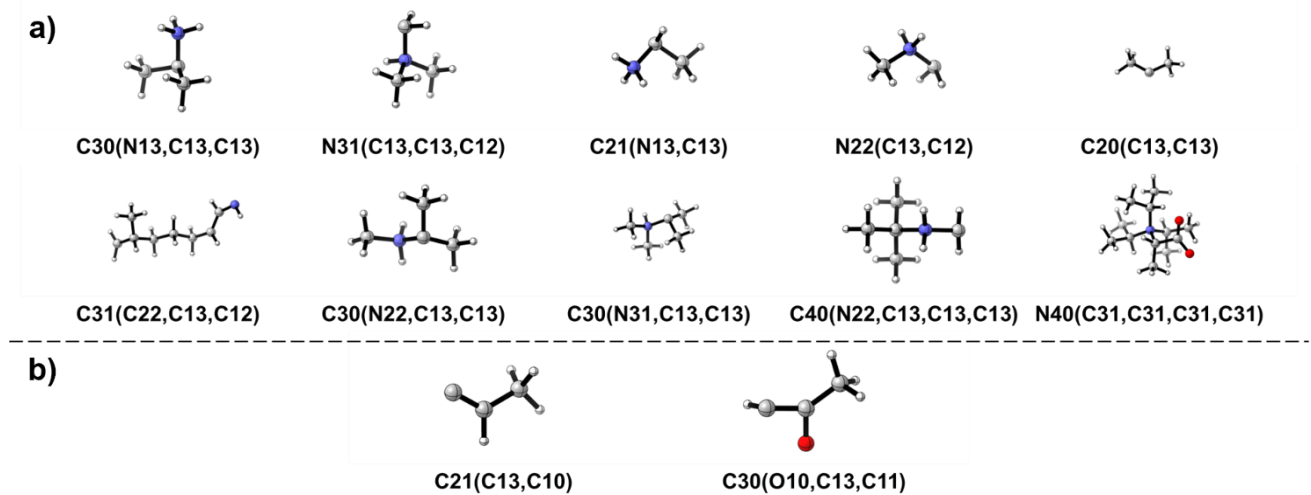

**Extended Data Fig. 1 Representative GCN1 types highlighting coverage differences across datasets.**

a) The ten most frequent GCN1 types present in FD25 but absent from PC3M, PC9, QM9 and ANI-1E, together with representative molecular structures.

b) Two GCN1 types observed in PC3M and PC9 but not included in FD25, with representative molecules.

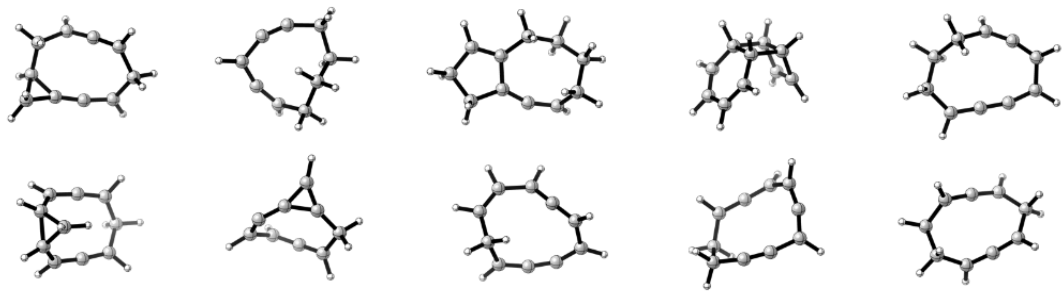

**Extended Data Fig. 2 Representative NBG0 types distinguishing FD25 from existing datasets.**

The ten most frequent NBG0 types present in FD25 but absent from PC3M, PC9, QM9 and ANI-1E, shown with representative molecular structures.

**Extended Data Table 2** Progressive refinement of GCN2 and NBG-Plus representations across levels.

| GCN2 level | NBG-Plus level | Step | Datapoints | MAE (kcal/mol) |
|---|---|---|---|---|
| 6 | 1 | 0 | 545018 | 10.43 |
| 7 | 2 | 1 | 807315 | 1.94 |
| 8 | 3 | 2 | 1255179 | 0.97 |
| 9 | 5 | 3 | 1744931 | 0.66 |
| 16 | 22 | 4 | 2233275 | 0.59 |

**Extended Data Table 3** Benchmark performance of GNN models on FD25 for ten quantum chemical properties.

| Target | MAE | Target | MAE |
|---|---|---|---|
| E | 0.7 | HOMO-LUMO Gap | 1.6 |
| H | 0.7 | alpha[a] | 0.2 |
| G | 0.7 | SMDE | $4.2 \times 10^{-2}$ |
| HOMO | 1.2 | Esolv | 0.1 |
| LUMO | 1.1 | Charge[a] | $3.7 \times 10^{-3}$ |

[a]In a.u.; all other quantities are in kcal/mol.

# Supporting Information

**1. Definition of symbols in Eq. 1 & Eq. 2**

$$\Psi_{VB} = \sum_k c_k \mathcal{A}\left[\phi_k^{(1)}(1)\phi_k^{(2)}(2)\cdots\phi_k^{(N_e)}(N_e)\chi_k\right] \tag{S1}$$

**Eq. S1** Modern valence-bond (VB) wavefunction, where: $\phi_k^{(i)}$ denotes the atomic or hybrid orbital (localized) occupied by the $i$-th electron in the $k$-th configuration; $\chi_k$ is the spin function associated with the $k$-th configuration (typically a singlet spin coupling for closed-shell systems); $\mathcal{A}$ is the antisymmetrization operator ensuring fermionic antisymmetry of the many-electron wavefunction; $c_k$ represents the weight coefficient of the $k$-th resonance structure, determined either by energy minimization or empirical parametrization.

$$E_{GCN0} = \frac{\sum_i o_i * e_i}{\sum_i o_i} \tag{S2}$$

**Eq. S2** Expression for $E_{GCN0}$, where $o_i$ denotes the occupancy of the $i$-th NBO orbital and $e_i$ denotes its energy.

## 2. Enumeration of GCN0

**Table S1** Enumeration of GCN0 (30 in total)

| Element | C | N | O | F |
|---|---|---|---|---|
| | C40 | N40 | O20 | F10 |
| | C30 | N30 | O10 | F01 |
| | C31 | N31 | O11 | |
| | C20 | N20 | O02 | |
| | C21 | N21 | | |
| | C22 | N22 | | |
| **GCN0** | C10 | N10 | | |
| | C11 | N11 | | |
| | C12 | N12 | | |
| | C13 | N13 | | |
| | C04 | N03 | | |
| | C02 | N01 | | |
| **Total** | 12 | 12 | 4 | 2 |

## 3. Algorithms

**Algorithm S1:** GCN1 List Generation

**Input:** gcn0_list ← list of GCN0
**Output:** gcn1_list ← list of GCN1
gcn0_conn_list ← empty list;
**for** gcn0 **in** gcn0_list **do**
    num_conns ← integer value of gcn0's second-last character;
    **if** num_conns ≠ 0 **then**
        append gcn0 to gcn0_conn_list;
    **end if**
**end for**
gcn1_list ← empty list;
**for** gcn0 **in** gcn0_list **do**
    num_conns ← integer value of gcn0's second-last character;
    **if** num_conns = 0 **then**
        continue;
    **end if**
    comb_list ← combinations_with_replacement(gcn0_conn_list, num_conns);
    **for** combo **in** comb_list **do**
        sorted_combo ← sort combo in reverse (descending) order;
        gcn1 ← gcn0 + "(" + join(sorted_combo, ",") + ")";
        append gcn1 to gcn1_list;
    **end for**
**end for**

Following Algorithm S1, enumeration starts from the 30 possible GCN0 types. Of these, 24 possess at least one non-hydrogen neighbor, yielding a gcn0_conn_list of length 24. Combinatorial expansion over these entries results in 47,864 unique GCN1 encodings.

For single–heavy-atom molecules (e.g., $CH_4$, $NH_3$), the only heavy atom has zero non-hydrogen connections, and thus its GCN1 is identical to its GCN0. Including these cases contributes 6 additional GCN1 types, leading to a total of 47,870 GCN1 encodings.

**Algorithm S2:** GCN2 List Generation

**Input:** gcn0_list ← list of GCN0
gcn1_list ← list of GCN1
**Output:** gcn2_list ← list of GCN2
regex ← compiled pattern of r'\((.*)\)';
gcn2_list ← empty list;
**for** gcn0 *in* gcn0_list **do**
    num_conns ← integer value of gcn0's second-last character;
    **if** num_conns = 0 **then**
        continue;
    **end if**
    gcn1_conn_list ← empty list;
    **for** gcn1 *in* gcn1_list **do**
        neighbors ← regex.search(gcn1).group(1);
        **if** gcn0 *is in* neighbors **then**
            append gcn1 to gcn1_conn_list;
        **end if**
    comb_list ← combinations_with_replacement(gcn1_conn_list, num_conns);
    **for** combo *in* comb_list **do**
        sorted_combo ← sort combo in reverse (descending) order;
        gcn2 ← gcn0 + "[" + join(sorted_combo, ",") + "]";
        append gcn2 to gcn2_list;
    **end for**
**end for**

Following Algorithm S2, when the input consists of the 13,302 GCN1 types recorded in FD25, the combinatorial expansion yields 1,067,046,693,717 theoretical GCN2 encodings. For single–heavy-atom molecules (e.g., $CH_4$, $NH_3$), the unique heavy atom has no non-hydrogen neighbors; thus, its GCN2 is identical to its GCN0. Including these cases contributes 6 additional GCN2 types, giving a total of 1,067,046,693,723 GCN2 encodings.

When the input instead uses the 47,870 GCN1 types obtained from Algorithm S1, the theoretical expansion produces 156,452,410,979,889 GCN2 encodings. Adding the same 6 single–heavy-atom cases results in a final total of 156,452,410,979,895 GCN2 encodings.

## 4. Data Standardization and Structural Normalization

PC3M was derived from the PubChemQC B3LYP dataset, which contains ~3M molecules optimized at the B3LYP/6-31G(d) level. In this study, we retained only neutral singlet structures composed of C/N/O/F non-hydrogen atoms and removed duplicates and connectivity-inconsistent entries based on their InChI identifiers.

QM9 is a subset of the GDB-17 chemical universe and contains 133,885 molecules with up to nine non-hydrogen atoms (C/N/O/F). All molecular geometries were optimized at the B3LYP/6-31G(2df,p) level, and multiple quantum-chemical properties, including total energy, enthalpy, and free energy, were computed at the same level. In this work, we removed 3,054 molecules that failed the standard consistency check, where Corina-generated Cartesian coordinates and the B3LYP-optimized geometries produce different SMILES strings (as documented in the QM9 README).

PC9 is a subset of the PubChemQC database containing 99,234 molecules with $N_a \leq 9$ (C/N/O/F non-hydrogen atoms). All geometries were optimized at the B3LYP/6-31G(d) level, and electronic energies were computed at the same level of theory. In this work, we removed 5,325 non-singlet species, retaining only neutral singlet molecules for analysis.

ANI-1E is a subset of the ANI-1 dataset comprising 57,462 molecules with $N_a \leq 8$ (C/N/O non-hydrogen atoms). Molecular geometries were obtained at the ωB97X-D/6-31G(d) level. We excluded 7 molecules with failed geometry optimizations, resulting in a cleaned set of 57,455 neutral species.

After the initial filtering, each structure was converted into an SDF file using Open Babel 2.4.0[1], enabling extraction of bond connectivity information. In addition, Open Babel 2.4.0 was used to generate InChI strings for all molecules, which were subsequently employed for structural comparison and consistency checks.

# 5. FD25 Coverage Analysis in Molecular Size, Binding Energy, GCN Encoding and Scaffold Topology

**5.1 Molecular Size (Na) and Binding Energy Distributions across Five Datasets**

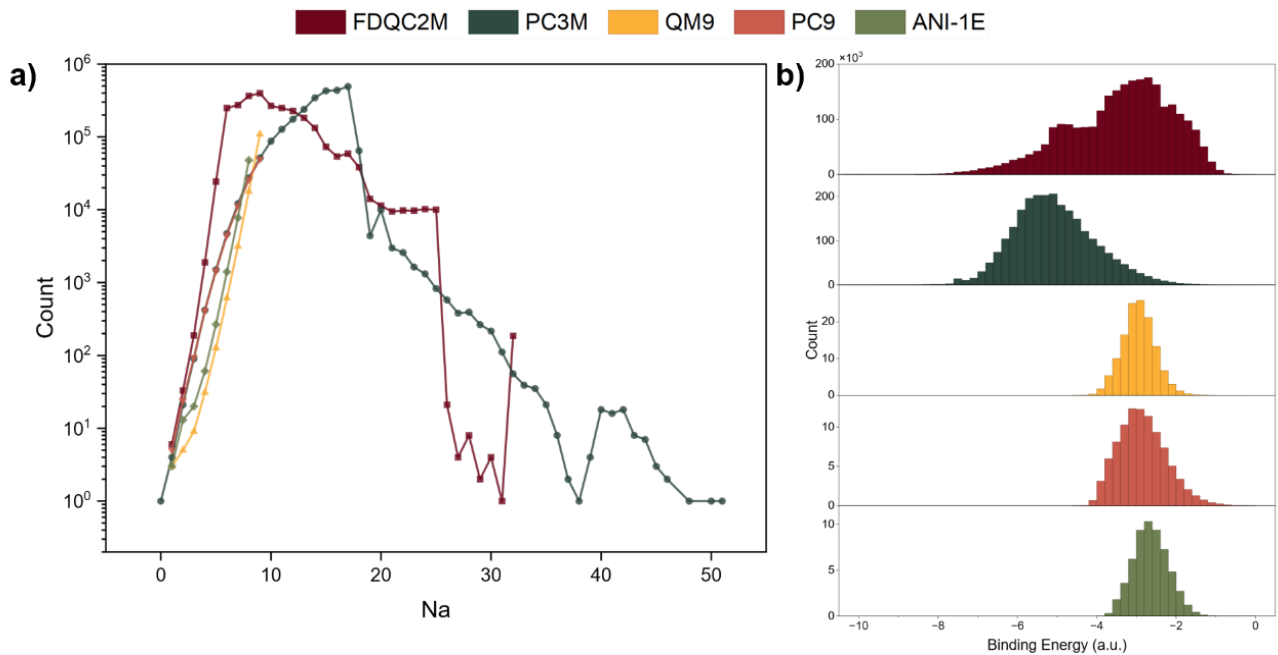

**Fig. S1** Molecular size (Na) and binding energy distributions across five datasets: FD25, PC3M, QM9, PC9, and ANI-1E.

a) Na Distributions.

FD25 provides dense and continuous coverage from small to medium-sized molecules (Na ≈ 6–20), bridging the gap between small-molecule datasets (QM9, PC9, ANI-1E; Na ≤ 9) and the broader but small-molecule-sparse PC3M dataset.

b) Binding energy distributions

Binding energy is calculated as the difference between the total energy of the molecule and the sum of the energies of its constituent isolated atoms (Eq. S3). Consistent with their Na coverage, FD25 exhibits the broadest and most structurally diverse binding energy distribution (≈ –8 to –1 a.u.), spanning multiple energy modes. PC3M shows a similarly wide but systematically lower distribution (≈ –7.5 to –2 a.u.). In contrast, QM9, PC9, and ANI-1E display narrow, small-molecule-restricted ranges (≈ –4 to –2 a.u.).

$$E_{bind} = E_{tot} - \sum_i N_i E_i \tag{S3}$$

## 5.2 KL Divergence Analysis of GCN Encoding and Scaffold

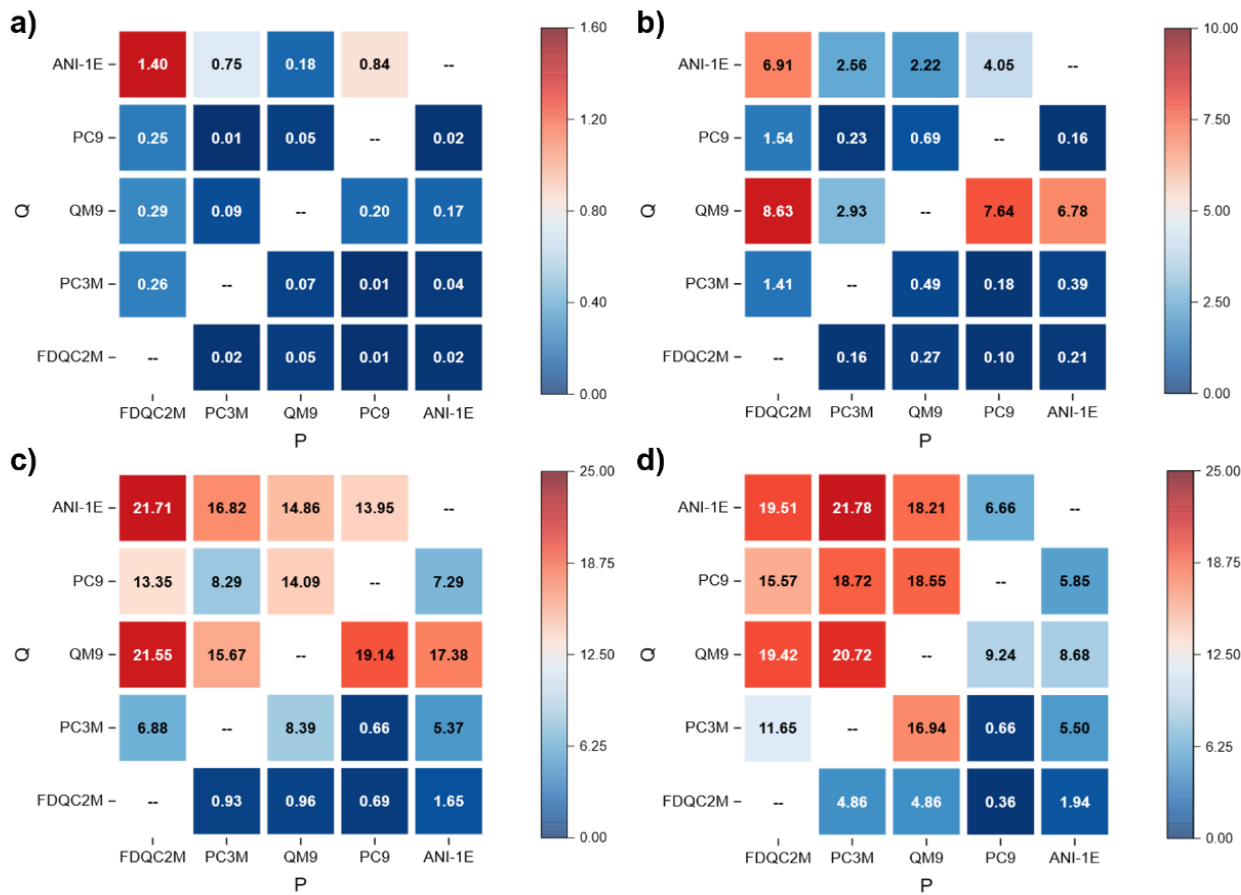

**Fig. S2** Heatmaps of Kullback–Leibler (KL) divergence for the distributions of a) GCN0, b) GCN1, c) GCN2, and d) Scaffold across five datasets.

KL divergence[2] is defined as

$$D_{KL}(P||Q) = \sum P(x) \log \frac{P(x)}{Q(x)}$$

As shown in Fig. S2, rows correspond to the reference distribution Q and columns to P. Lower values indicate closer similarity. For GCN0, the information is relatively simple, and thus the overall $D_{KL}(P||Q)$ values are close to 0. Notably, in the ANI-1E dataset, there is no F element, leading to a deficiency of two types of GCN0 related to F, which impairs its ability to fit GCN0 distributions from other datasets. For GCN1, the information becomes richer. Both QM9 and ANI-1E show limited capability in fitting the GCN1 distributions of other datasets, whereas PC3M and PC9 maintain a relatively high performance level. With GCN2, as the information complexity increases further, the fitting capabilities of PC3M, QM9, PC9, and ANI-1E deteriorate significantly, with only FD25 retaining a high level of performance.

For scaffold, FD25 shows the lowest KL divergence when serving as the reference, meaning its scaffold distribution can effectively approximate those of other datasets, whereas the reverse is not true. This highlights the broader and more balanced scaffold coverage of FD25.

# 6. GNN Architecture, Hyperparameter Tuning, and Training Strategy

The architecture and hyperparameter settings of the GNN model employed in this study were adapted from previously reported studies[3] and pre-optimized on the QM9 dataset to ensure reasonable performance on typical small-molecule datasets.

## 6.1 GNN Architecture

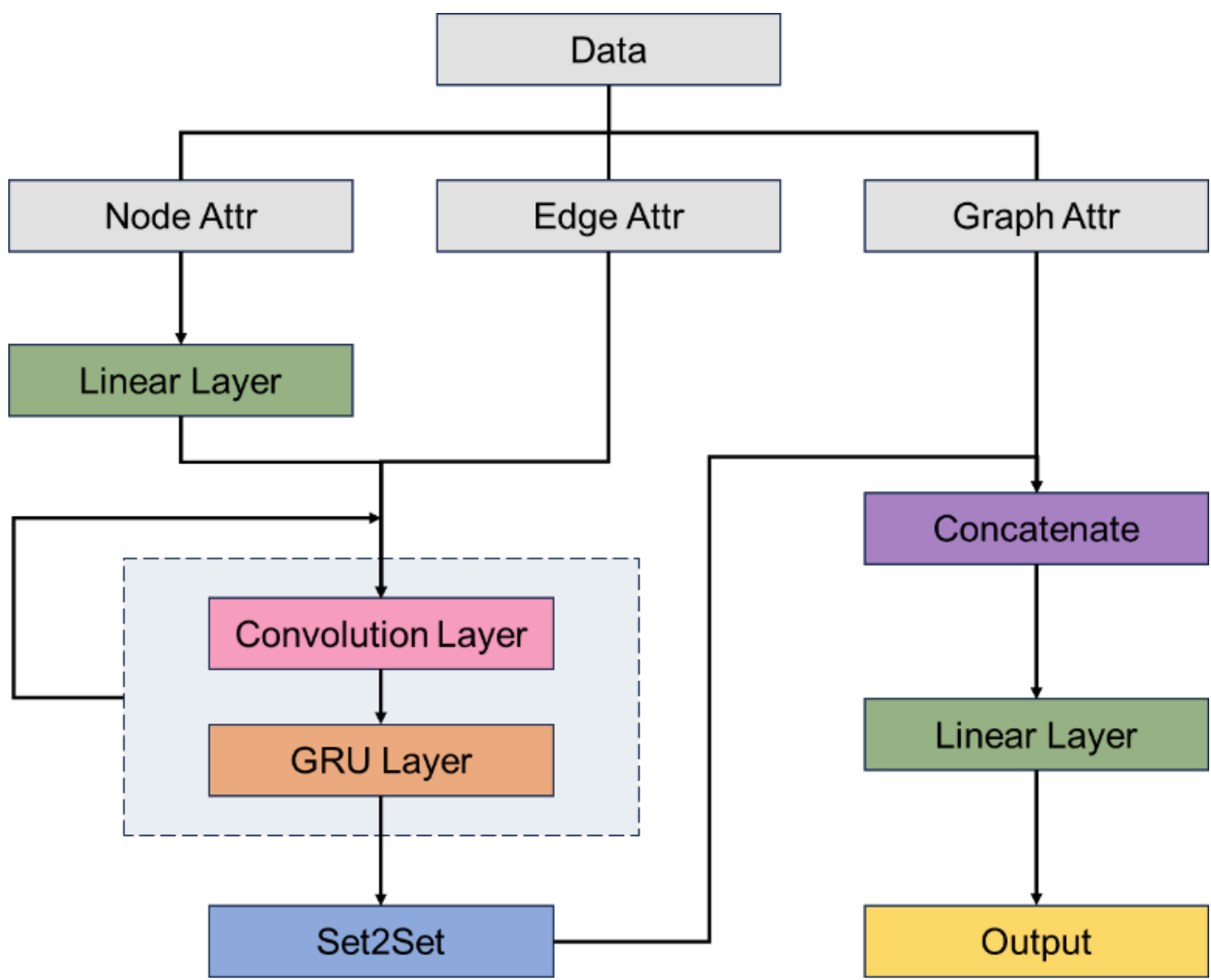

**Fig. S3** Detailed structure of the GNN model.

It should be noted that only node and edge features are employed in this work; graph-level features are not utilized. Molecular graphs are constructed by representing atoms as nodes and chemical bonds as edges to capture molecular structure. Node features encode various atomic properties, such as element type, formal charge, and number of bonded hydrogens, while edge features convey bond-related information, including bond type and bond length. The specific node and edge feature descriptors used in this study are summarized in Tables S1 and S2, respectively.

Importantly, to avoid potential ambiguities in bond order assignment, all bond orders in the input structures were uniformly set to 1 prior to model training. Consequently, the "edge type" feature in this work serves solely to indicate the presence or absence of a bond, rather than to distinguish between single, double, or other bond orders.

**Table S2** Node attributes used in this model

| Feature | Description | Type | Range |
|---|---|---|---|
| Element type | One-hot encoded element type | vector of int | {0, 1} |
| Atomic number | Atomic number | int | {1, 6, 7, 8, 9} |
| Num neighbors | Number of neighboring atoms | int | [1, 4] |

| | Num hs | Number of neighboring hydrogens | int | [0, 4] |
|---|---|---|---|---|

Given that the objective of this study is to predict scalar properties of molecules such as energy (E), enthalpy (H), and Gibbs free energy (G), it is essential that the model's output remains invariant to structural transformations, including rotations and translations. To achieve this without introducing complex equivariant neural networks, we opted to use inter-node distances as edge attributes. Specifically, we used $\frac{1}{r}$ rather than the direct distance $r$ to compress the range of large distances, thereby reducing the model's sensitivity to minor variations at longer ranges while retaining the significance of changes at shorter distances. This approach is particularly crucial for training on larger molecules.

**Table S3** Edge attributes used in this model

| Feature | Description | Type | Range |
|---|---|---|---|
| Edge type | One-hot encoded bond type | vector of int | {0, 1} |
| 1/r | The reciprocal of the distance between nodes | float | (0, +∞] |

## 6.2. Hyperparameter Tuning

To identify the optimal hyperparameter configuration, we conducted hyperparameter tuning on the QM9 dataset. Specifically, Bayesian optimization was employed to evaluate 100 hyperparameter combinations within the ranges specified in Table S4.

The target properties were standardized using Z-score normalization, and the mean absolute error (MAE) was used as the loss function. All models were trained using the Adam optimizer with a batch size of 64 for 100 epochs. During training, the learning rate was dynamically adjusted via the ReduceLROnPlateau scheduler: if the validation loss failed to decrease for 20 consecutive epochs, the learning rate was multiplied by a decay factor of 0.7. Model selection and early stopping were based on performance on the validation set, and final results are reported on the held-out test set.

Given that models of varying complexity exhibit different convergence behaviors, we selected the top-performing candidates from the initial screening and extended their training to 500 epochs. The results show that, after sufficient training, their predictive performances converge to nearly identical levels. Since the primary objective of this work is not to achieve state-of-the-art accuracy but rather to ensure prediction errors below the chemical accuracy threshold (1 kcal/mol), we opted for a relatively simpler model architecture to reduce computational cost and training time.

**Table S4** Hyperparameter search using Bayesian Optimization

| Parameters | Range | Selected Value |
|---|---|---|
| Learning rate | [0.0001, 0.01] | 0.001 |
| Message passing times | {3, 4, 5, 6, 7, 8} | 3 |
| Processing steps in set2set | {1, 2, 3, 4, 5, 6} | 3 |
| Dimension of linear layer | {32, 48, 64, 80, 96, 112, 128} | 64 |
| Dimension of conv layer | {32, 48, 64, 80, 96, 112, 128} | 64 |

## 6.3 Training Strategy

We trained and evaluated our models on the FD25, PC3M, QM9, PC9, and ANI-1E datasets, using the mean absolute error (MAE) and root mean square deviation (RMSD) of binding energies as the primary evaluation metrics. Two data-splitting strategies were employed: random splitting and manual splitting, the latter dividing each dataset into a core set and an extended set.

Random splitting was applied to QM9, PC9, and ANI-1E. For QM9, 10,000 samples were randomly selected as the validation set and another 10,000 as the test set, with the remainder used for training. For PC9 and ANI-1E, the datasets were randomly partitioned into training, validation, and test sets in an 8:1:1 ratio.

In the manual splitting scheme, the core sets of PC3M, QM9, PC9, and ANI-1E all include: all GCN1 and GCN2 molecules, all NBG0 and NBG-Plus molecules, and the element-composition-balanced subset.

For FD25, the core set further includes: the comprehensive conformational/configurational subset for systems with $Na \leq 6$, and the conformational space for systems with $6 < Na \leq 8$.

The extended set (i.e., all samples outside the core set) was randomly divided into three equal parts (1:1:1). One part was combined with the core set to form the training set, while the other two served as the validation and test sets, respectively.

Notably, due to the small size of the PC3M core set—accounting for only ~1/6 of the entire dataset—the extended set was instead split in a 4:1:1 ratio. The largest portion (4/6) was merged with the core set for training, while the remaining two portions were used for validation and testing.

Evaluation on QM9, PC9, and ANI-1E demonstrates that this manual splitting strategy yields significantly better generalization performance compared to random splitting. Consequently, all results reported in this work are based on the manual splitting protocol.

**Table S5** Test MAE for models trained on random and manual data splits

|          | Test MAE (kcal/mol) | |
| --- | --- | --- |
|          | Random | Manual |
| **FDQC2M** | /      | 0.5883 |
| **PC3M**   | /      | 0.5245 |
| **QM9**    | 0.6082 | **0.5129** |
| **PC9**    | 1.2025 | **0.9137** |
| **ANI-1E** | 0.9086 | **0.7306** |

## 7. Level Definitions of Structural Expansion and Error Attribution Analysis

**Table S6** Number of structures with different levels of GCN2 and NBG-Plus in FD25

| GCN2 level | Number | NBG-Plus level | Number |
|---|---|---|---|
| 0 | 6 | 0 | 57922 |
| 1 | 33 | 1 | 158018 |
| 2 | 146 | 2 | 253192 |
| 3 | 1346 | 3 | 304157 |
| 4 | 133893 | 4 | 209457 |
| 5 | 258493 | 5 | 153941 |
| 6 | 396380 | 6 | 100440 |
| 7 | 471562 | 7 | 66852 |
| 8 | 454006 | 8 | 43635 |
| 9 | 279135 | 9 | 26407 |
| 10 | 135814 | 10 | 15713 |
| 11 | 64572 | 11 | 11688 |
| 12 | 28843 | 12 | 9746 |
| 13 | 6920 | 13 | 6165 |
| 14 | 1708 | 14 | 4375 |
| 15 | 338 | 15 | 3326 |
| 16 | 80 | 16 | 2369 |
| | | 17 | 1829 |
| | | 18 | 1781 |
| | | 19 | 1379 |
| | | 20 | 285 |
| | | 21 | 28 |
| | | 22 | 12 |

Definition of GCN2 level: level-n corresponds to the structures containing n+1 non-hydrogen atoms in the GCN2 encoding.
Definition of NBG-Plus level: level-0 corresponds to ring/cage structures derived from NBG0 via isovalent substitution of carbon atoms with heteroatoms. For n≥1, level-n denotes structures satisfying $\max(\mathcal{V}_c, \mathcal{E}_c) = n$.

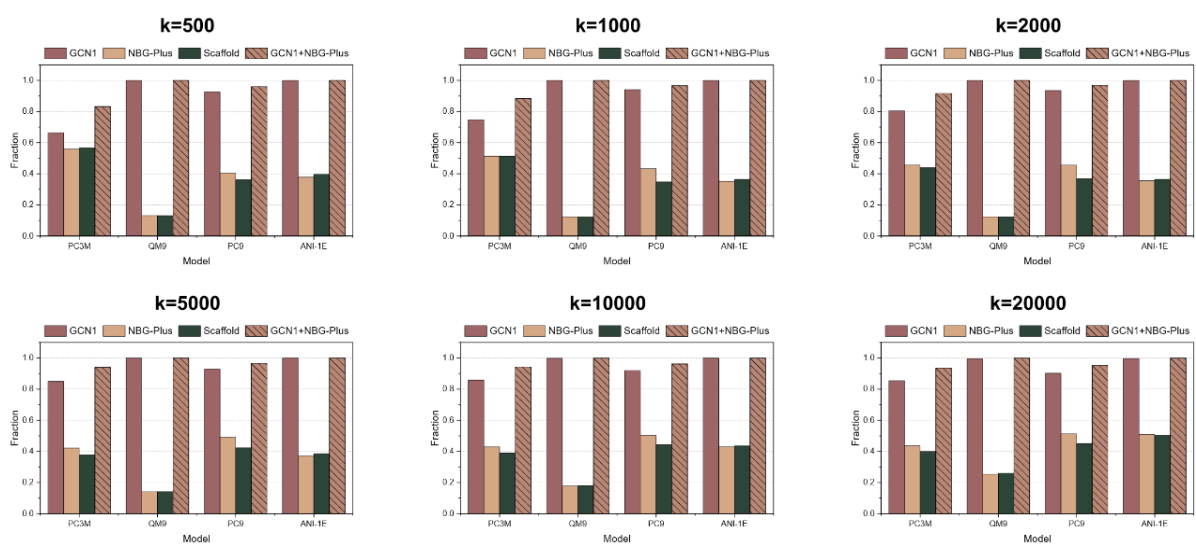

**Fig. S4** Top-k error attribution with different k of 500, 1000, 2000, 5000, 10000 and 20000.